\documentclass{article}
\usepackage{spconf,amsmath,graphicx}
\usepackage{booktabs}
\usepackage{multirow}
\usepackage{float}
\usepackage[colorlinks,
            linkcolor=blue,       
            anchorcolor=blue,  
            citecolor=blue,        
            ]{hyperref}

\title{CPTR: Full Transformer Network for Image Captioning}
\name{Wei Liu\thanks{* Wei Liu and Sihan Chen contribute equally to this paper. }\textsuperscript{\rm{*1,2}}, Sihan Chen\textsuperscript{\rm{*1,2}}, Longteng Guo\textsuperscript{\rm{1,2}}, Xinxin Zhu\textsuperscript{\rm{1}}, Jing Liu\textsuperscript{\rm{1,2}}}
\address{\textsuperscript{1} National Laboratory of Pattern Recognition, Institute of Automation, Chinese Academy of Sciences \\
   \textsuperscript{2} School of Artificial Intelligence, University of Chinese Academy of Sciences}
\begin{document}
%
\maketitle

\begin{abstract}
In this paper, we consider the image captioning task from a new sequence-to-sequence prediction perspective and propose CaPtion TransformeR (CPTR) which takes the sequentialized raw images as the input to Transformer. Compared to the ``CNN+Transformer" design paradigm, our model can model global context at every encoder layer from the beginning and is totally convolution-free. Extensive experiments demonstrate the effectiveness of the proposed model and we surpass the conventional ``CNN+Transformer" methods on the MSCOCO dataset. Besides, we provide detailed visualizations of the self-attention between patches in the encoder and the ``words-to-patches" attention in the decoder thanks to the full Transformer architecture.   
\end{abstract}
\section{Introduction}
\label{sec:intro}

Image captioning is a challenging task which concerns about generating a natural language to describe the input image automatically. Currently, most captioning algorithms follow an encoder-decoder architecture in which a decoder network is used to predict words according to the feature extracted by the encoder network via attention mechanism. Inspired by the great success of Transformer \cite{vaswani2017attention} in the natural language processing field, recent captioning models tend to replace the RNN model with Transformer in the decoder part for its capacity of parallel training and excellent performance, however, the encoder part always remains unchanged, i.e., utilizing a CNN model (e.g. ResNet \cite{he2016deep}) pretrained on image classification task to extract spatial feature or a Faster-RCNN \cite{ren2016faster} pretrained on object detection task to extract bottom-up \cite{anderson2018bottom} feature. 

Recently, researches about applying Transformer to computer vision field have attracted extensive attention. For example, DETR \cite{carion2020end} utilizes Transformer to decode detection predictions without prior knowledge such as region proposals and non-maximal suppression. ViT \cite{dosoViTskiy2020image} firstly utilizes Transformer without any applications of convolution operation for image classification and shows promising performance especially when pretrained on very huge datasets (i.e., ImageNet-21K, JFT). After that, full Transformer methods for both high-level and low-level down-stream tasks emerge, such as SETR \cite{zheng2020rethinking} for image semantic segmentation and  IPT \cite{chen2020pre} for image processing. 

Inspired by the above works, we consider solving the image captioning task from a new sequence-to-sequence perspective and propose CaPtion TransformeR (CPTR), a full Transformer network to replace the CNN in the encoder part with Transformer encoder which is totally convolution-free. Compared to the conventional captiong models taking as input the feature extracted by CNN or object detector,  we directly sequentialize raw images as input. Specifically, we divide an image into small patches of fixed size (e.g. $16\times16$), flatten each patch and reshape them into a 1D patch sequence. The patch sequence passes through a patch embedding layer and a learnable positional embedding layer before being fed into the Transformer encoder.

Compared to the ``CNN+Transformer" paradigm, CPTR is a more simple yet effective method that totally avoids convolution operation. Due to the local operator essence of convolution, the CNN encoder has limitation in global context modeling which can only  be fulfilled by enlarging receptive field gradually as the convolution layers go deeper. However, encoder of CPTR can utilize long-range dependencies among the sequentialized patches from the very beginning via self-attention mechanism. During the generation of words, CPTR models ``words-to-patches" attention in the cross attention layer of decoder which is proved to be effective. We evaluate our method on MSCOCO image captioning dataset and it outperforms both ``CNN+RNN" and ``CNN+Transformer" captioning models.

\begin{figure*}[t]
\begin{center}

\includegraphics[width=0.85\linewidth]{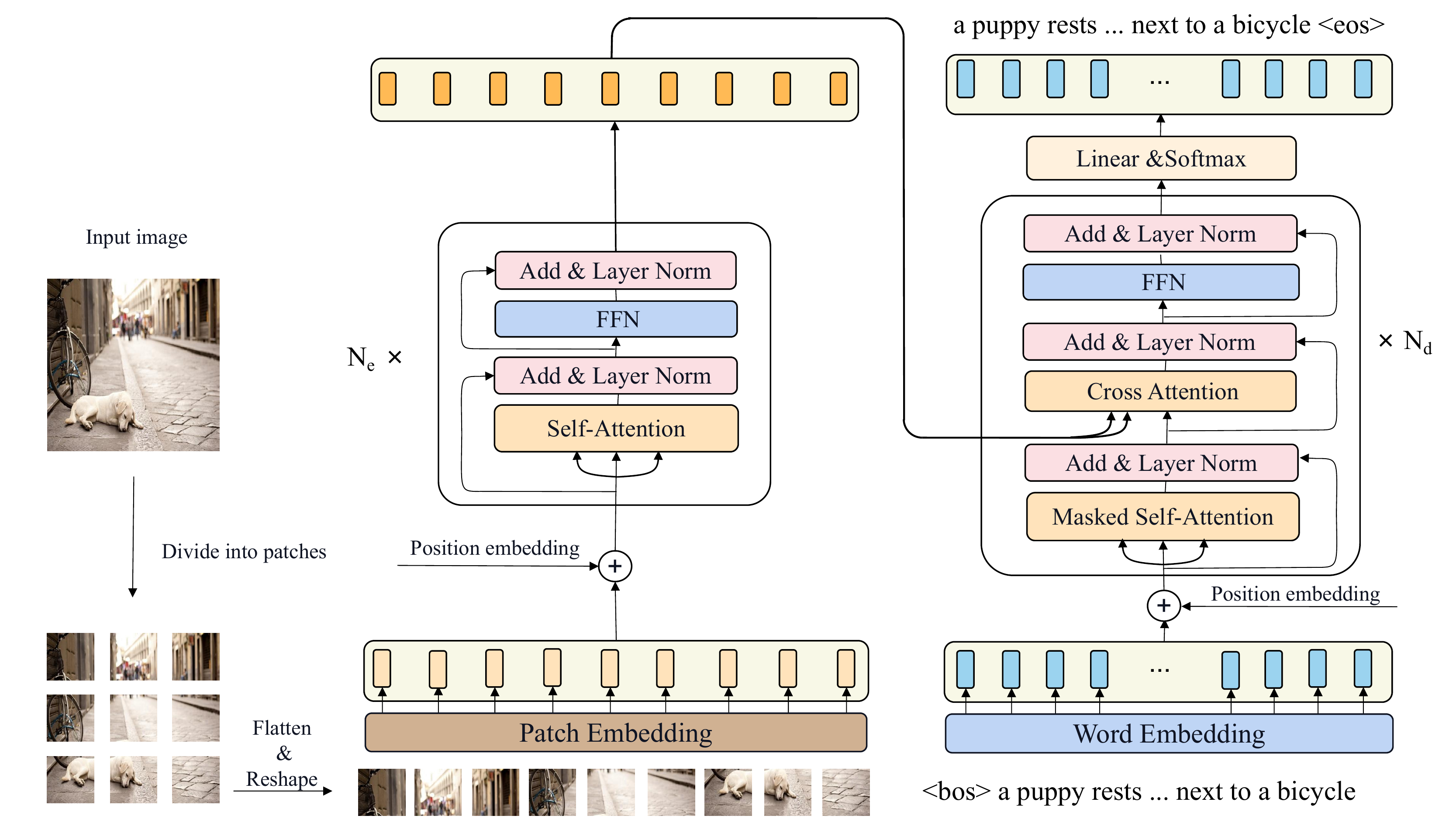}
\end{center}
\vspace{-0.5cm}
\caption{ The overall architecture of proposed CPTR model.}
\vspace{-0.4cm}
\label{fig:model}
\end{figure*}

\section{FrameWork}
\label{sec:pagestyle}
\subsection{Encoder}
As depicted in Fig. \ref{fig:model}, instead of using a pretrained CNN or Faster R-CNN model to extract spatial features or bottom-up features like the previous methods, we choose to sequentialize the input image and treat image captioning as a sequence-to-sequence prediction task. Concretely, we divide the original image into a sequence of image patches to adapt to the input form of Transformer.

Firstly, we resize the input image into a fixed resolution $X\in R^{H\times W\times 3}$ (with 3 color channels), then divide the resized image into N patches, where ${N = \frac{H}{P} \times \frac{W}{P}}$  and $P$ is the patch size ($P=16$ in our experiment settings). After that, we flatten each patch and reshape them into a 1D patch sequence $ X_p\in R^{N\times (P^{2} \cdot 3)}$. We use a linear embedding layer to map the flattened patch sequence to latent space and add a learnable 1D position embedding to the patch features, then we get the final input to the Transformer encoder which is denoted as $P_a = \left[p_{1}, \ldots, p_{N}\right]$. 

The encoder of CPTR consists of $N_e$ stacked identical layers, each of which consists of a multi-head self-attention (MHA) sublayer followed by a positional feed-forward sublayer. MHA contains $H$ parallel heads and each head $h_i$ corresponds to an independent scaled dot-product attention function which allows the model to jointly attend to different subspaces. Then a linear transformation $W^O$ is used to aggregate the attention results of different heads, the process can be formulated as follows: 
\begin{equation}
\operatorname{MHA}(Q, K, V)=\operatorname{Concat}\left(h_{1}, \ldots, h_{H}\right) W^{O}
\end{equation}
The scaled dot-product attention is a particular attention proposed in Transformer model, which can be computed as follows:
\begin{equation}
\operatorname{Attention}(Q, K, V)=\operatorname{Softmax}\left(\frac{Q K^{T}}{\sqrt{d_k}}\right) V 
\end{equation}
where \textit Q $\in{R^{N_q \times \textit d_k}}$, \textit K $\in{R^{N_k \times \textit d_k}}$ and \textit V $\in{R^{N_k \times \textit d_v}}$ are the query, key and value matrix, respectively.

The followed positional feed-forward sublayer is implemented as two linear layers with GELU activation function and dropout between them to further transform features. It can be formulated as:
\begin{equation}
\mathrm{FFN}(x)=\mathrm{FC_2}(\operatorname{Dropout}(\operatorname{GELU}(\mathrm{FC_1}(x))))
\end{equation}

In each sublayer, there exists a sublayer connection composed of a residual connection, followed by layer normalization.
\begin{equation}
     x^{out}=\operatorname{LayerNorm}(x^{in} + \text{Sublayer}(x^{in})))
\end{equation}

where $x^{in}$, $x^{out}$ are the input and output of one sublayer respectively and the sublayer can be attention layer or feed forward layer.

\begin{table*}[t]
\begin{tabular}{@{}l l l l l l l l l l l l l l l@{}}
\toprule
\multicolumn{1}{c}{\multirow{2}{*}{Model}} & \multicolumn{2}{c}{BLEU-1}                           & \multicolumn{2}{c}{BLEU-2}                           & \multicolumn{2}{c}{BLEU-3}                           & \multicolumn{2}{c}{BLEU-4}                           & \multicolumn{2}{c}{METEOR}                             & \multicolumn{2}{c}{ROUGE}                             & \multicolumn{2}{c}{CIDEr}                            \\ \cmidrule(l){2-15} 
\multicolumn{1}{c}{}                       & \multicolumn{1}{c}{c5} & \multicolumn{1}{c}{c40} & \multicolumn{1}{c}{c5} & \multicolumn{1}{c}{c40} & \multicolumn{1}{c}{c5} & \multicolumn{1}{c}{c40} & \multicolumn{1}{c}{c5} & \multicolumn{1}{c}{c40} & \multicolumn{1}{c}{c5} & \multicolumn{1}{c}{c40} & \multicolumn{1}{c}{c5} & \multicolumn{1}{c}{c40} & \multicolumn{1}{c}{c5} & \multicolumn{1}{c}{c40} \\ 
\midrule

         \textbf{CNN+RNN}\\
\midrule
SCST \cite{rennie2017self}                                        & 78.1                    & 93.7                     & 61.9                    & 86.0                     & 47.0                    & 75.9                     & 35.2                    & 64.5                     & 27.0                    & 35.5                     & 56.3                    & 70.7                     & 114.7                   & 116.0                   \\
LSTM-A \cite{yao2017boosting}& 78.7                    & 93.7                     & 62.7                    & 86.7                     & 47.6                    & 76.5                     & 35.6                    & 65.2                     & 27.0                    & 35.4                     & 56.4                    & 70.5                     & 116.0                   & 118.0                   \\
Up-Down \cite{anderson2018bottom}                                     & 80.2                    & 95.2                     & 64.1                    & 88.8                     & 49.1                    & 79.4                     & 36.9                    & 68.5                     & 27.6                    & 36.7                     & 57.1                    & 72.4                     & 117.9                   & 120.5                   \\
RF-Net \cite{ke2019reflective}                                      & 80.4                    & 95.0                     & 64.9                    & 89.3                     & 50.1                    & 80.1                     & 38.0                    & 69.2                     & 28.2                    & 37.2                     & 58.2                    & 73.1                     & 122.9                   & 125.1                   \\
GCN-LSTM \cite{yao2018exploring}                                    & -                       & -                        & 65.5                    & 89.3                     & 50.8                    & 80.3                     & 38.7                    & 69.7                     & 28.5                    & 37.6                     & 58.5                    & 73.4                     & 125.3                   & 126.5                   \\
SGAE \cite{yang2019auto}                                        & 81.0                    & 95.3                     & 65.6                    & 89.5                     & 50.7                    & 80.4                     & 38.5                    & 69.7                     & 28.2                    & 37.2                     & 58.6                    & 73.6                     & 123.8                   & 126.5                   \\

\midrule

         \textbf{CNN+Transformer}\\
\midrule
ETA \cite{li2019entangled}  &81.2& 95.0& 65.5 &89.0 & 50.9&80.4&38.9 &70.2 &28.6 &38.0 &58.6&73.9&122.1 & 124.4 \\

\midrule
CPTR  &\textbf{81.8} &95.0 &\textbf{66.5} &89.4  & \textbf{51.8} &\textbf{80.9} &\textbf{39.5} &\textbf{70.8} &\textbf{29.1} &\textbf{38.3} &\textbf{59.2} &\textbf{74.4} &\textbf{125.4} &\textbf{127.3}           
\\ \bottomrule
\end{tabular}
 \caption{Performance comparisons on MSCOCO online test server. All models are finetuned with self-critical training. c5/c40 denotes the official test settings with 5/40 ground-truth captions.  }
    \label{tab:COCO_online}
    \vspace{-0.4cm}
\end{table*}

\subsection{Decoder}

In the decoder side, we add sinusoid positional embedding to the word embedding features and take both the addition results and encoder output features as the input. 
The decoder consists of $N_d$ stacked identical layers with each layer containing a masked multi-head self-attention sublayer  followed by a multi-head cross attention sublayer and a positional feed-forward sublayer sequentially. 

The output feature of the last decoder layer is utilized to predict next word via a linear layer whose output dimension equals to the vocabulary size. Given a ground truth sentence $y^*_{1:T}$ and the prediction $y^*_{t}$ of captioning model with parameters $\theta$, we minimize the following cross entropy loss:
\begin{equation}
\mathrm L_{X E}(\theta)=-\sum_{t=1}^{T} \log \left(p_{\theta}\left(y_{t}^{*} \mid y_{1: t-1}^{*}\right)\right)
\end{equation}
Like other captioning methods, we also finetune our model using self-critical training \cite{rennie2017self}.
\section{Experiments}
\subsection{Dataset and Implementation Details}
We evaluate our proposed model on MS COCO \cite{lin2014microsoft} dataset which is the most commonly used benchmark for image captioning. To be consistent with previous works, we use the ``Karpathy splits" \cite{karpathy2015deep} which contains 113,287, 5,000 and 5,000 images for training, validation and test, respectively. The results are reported on both the Karpathy test split for offline evaluation and MS COCO test server for online evaluation.  

We train our model in an end-to-end fashion with the encoder initialized by the pre-trained ViT model. The input images are resized to  $384\times384$ resolution and the patch size is setting to 16. The encoder contains 12 layers and decoder contains  4 layers. Feature dimension is 768, and the attention head number is 12 for both encoder and decoder. The whole model is first trained with cross-entropy loss for 9 epochs using an initial learning rate of $3\times 10^{-5}$ and decayed by 0.5 at the last two epochs. After that, we finetune the model using self-critical training \cite{rennie2017self}  for 4 epochs with an initial learning rate of $7.5\times 10^{-6}$ and decayed by 0.5 after 2 epochs. We use Adam optimizer and  the batch size is 40. Beam search is used and the beam size is 3.

We use  BLEU-1,2,3,4, METEOR, ROUGE and CIDEr scores \cite{chen2015microsoft} to evaluate our method which are denoted as B-1,2,3,4, M, R and C, respectively.

\begin{table}[htbp]
    \centering
    \scalebox{0.8}{
    \begin{tabular}{l c c c c c c c}
    \toprule
        Method & B-1 &B-2& B-3 & B-4&  M &R&C \\
    \midrule
         \textbf{CNN+RNN}\\
    \midrule
        LSTM \cite{vinyals2015show}  &-& -& - &31.9 & 25.5 & 54.3&106.3 \\
        SCST \cite{rennie2017self}  &-& -& - &34.2 & 26.7 & 55.7&114.0 \\
        LSTM-A \cite{yao2017boosting}  &78.6& -& - &35.5 & 27.3 & 56.8&118.3 \\
        RFNet \cite{ke2019reflective}  &79.1& 63.1& 48.4 &36.5 & 27.7 & 57.3&121.9 \\
        Up-Down \cite{anderson2018bottom}  &79.8& -& - &36.3 & 27.7 & 56.9&120.1 \\
        GCN-LSTM \cite{yao2018exploring}  &80.5& -& - &38.2 & 28.5 & 58.3&127.6 \\
        LBPF \cite{qin2019look}  &80.5& -& - &38.3 & 28.5 & 58.4&127.6 \\
        SGAE \cite{yang2019auto}  &80.8& -& - &38.4 & 28.4 & 58.6&127.8 \\
    \midrule
         \textbf{CNN+Transformer}\\
    \midrule
        ORT \cite{herdade2019image}  &80.5& -& - &38.6 & 28.7 & 58.4&128.3 \\
        ETA \cite{li2019entangled}  &81.5& -& - &39.3 & 28.8 & 58.9&126.6 \\
    \midrule
        CPTR   &\textbf{81.7}&\textbf {66.6}& \textbf{52.2} &\textbf{40.0} & \textbf{29.1} &\textbf{59.4} &\textbf{129.4} \\
    \bottomrule
        
    \end{tabular}}
   
    \caption{Performance comparisons on COCO Karpathy test split. All models are finetuned with self-critical training. }
    \label{tab:kaparthy test}
\end{table}

\subsection{Performance Comparison}
We compare proposed CPTR to ``CNN+RNN" models including LSTM \cite{vinyals2015show}, SCST \cite{rennie2017self}, LSTM-A \cite{yao2017boosting}, RFNet \cite{ke2019reflective}, Up-Down \cite{anderson2018bottom}, GCN-LSTM \cite{yao2018exploring}, LBPF \cite{qin2019look}, SGAE \cite{yang2019auto} and ``CNN+Transformer" models including ORT \cite{herdade2019image}, ETA \cite{li2019entangled}. These methods mentioned above all use image features extract by a CNN or object detector as inputs, while our model directly takes the raw image as input. Table \ref{tab:kaparthy test} shows the performance comparison results on  the offline Karpathy test split, and CPTR achieves 129.4 Cider score which outperforms both ``CNN+RNN" and ``CNN+Transformer" models. We attribute the superiority of CPTR model over conventional ``CNN+" architecture to the capacity of modeling global context at all encoder layers. The online COCO test server evaluation results shown in Table \ref{tab:COCO_online} also demonstrates the effectiveness of our CPTR model.

\subsection{Ablation Study}
We conduct ablation studies from the following aspects: (a) Different pre-trained models to initialize the Transformer encoder. (b) Different resolutions of input image. (c) The number of layers and feature dimension in the Transformer decoder. All experiments are conducted on the Karpathy validation set and optimized by cross-entropy loss only. 

The experiment results are shown in Table \ref{tab:abalation_study} from which we can draw the following conclusions. Firstly, pretraining vitals for CPTR model. Compared to training from scratch, using parameters of the ViT model pretrained on ImageNet-21K dataset to initialize CPTR encoder brings significant performance gains. Besides, using the parameters of the ViT model finetuned on the ImageNet 2012 dataset to initialize the encoder further brings one point improvement on the CIDEr score. Secondly, CPTR is little sensitive to the decoder hyper-parameter  including the number of layers and feature dimension, among which 4 layers, 768 dimensions shows the best performance (111.6 Cider score). Regards to the input image resolution, we found that increasing it from $224\times224$  to $384\times384$  while maintaining the patch size equals to 16 can bring huge performance gains (from 111.6 Cider score to 116.5 Cider score). It is sensible for that the length of patch sequence increases from 196 to 576 due to the increasing input resolution, and can divide image more specifically and provide more features to interact with each other via the encoder self-attention layer.

\begin{table}[t]
    \centering
    \scalebox{0.8}{
    \begin{tabular}{c c c c c c c c}
    \toprule
        Pretrained Model&Res& \#Layer& Dim& B-4&  M &R&C \\
    \midrule
        From scratch &224& 4& 768 & 16.5 &17.3 &   42.1&45.5 \\    
        ViT21K &224& 4& 768 &33.5 &27.4 &   55.8&110.6 \\
        
        ViT21K+2012 &224& 4& 768 & 33.8 &27.4 &   55.8&111.6 \\

        ViT21K+2012 &224& 1& 768 & 33.8 &27.5 &   56.1&111.2 \\

        ViT21K+2012 &224& 2& 768 & 33.4 &27.5 &   56.0&110.9 \\
        ViT21K+2012 &224& 6& 768 &33.7 &27.5 &   55.9&110.9 \\
        ViT21K+2012 &224& 4& 512 &34.0 &27.4 &   56.0&111.0 \\

        ViT21K+2012 &384& 4& 768 & \textbf{34.9} &\textbf{28.2} &\textbf{   56.9}&\textbf{116.5} \\
          
    \bottomrule
        
    \end{tabular}}
    \caption{ Ablation studies on the cross-entropy training stage. Res: image resolution. \#Layer: the number of decoder layers. Dim: the feature dimension of decoder.  }
    \vspace{-0.3cm}
    \label{tab:abalation_study}
    
\end{table}
\begin{figure}[h]
\begin{center}
\includegraphics[width=1.0\linewidth]{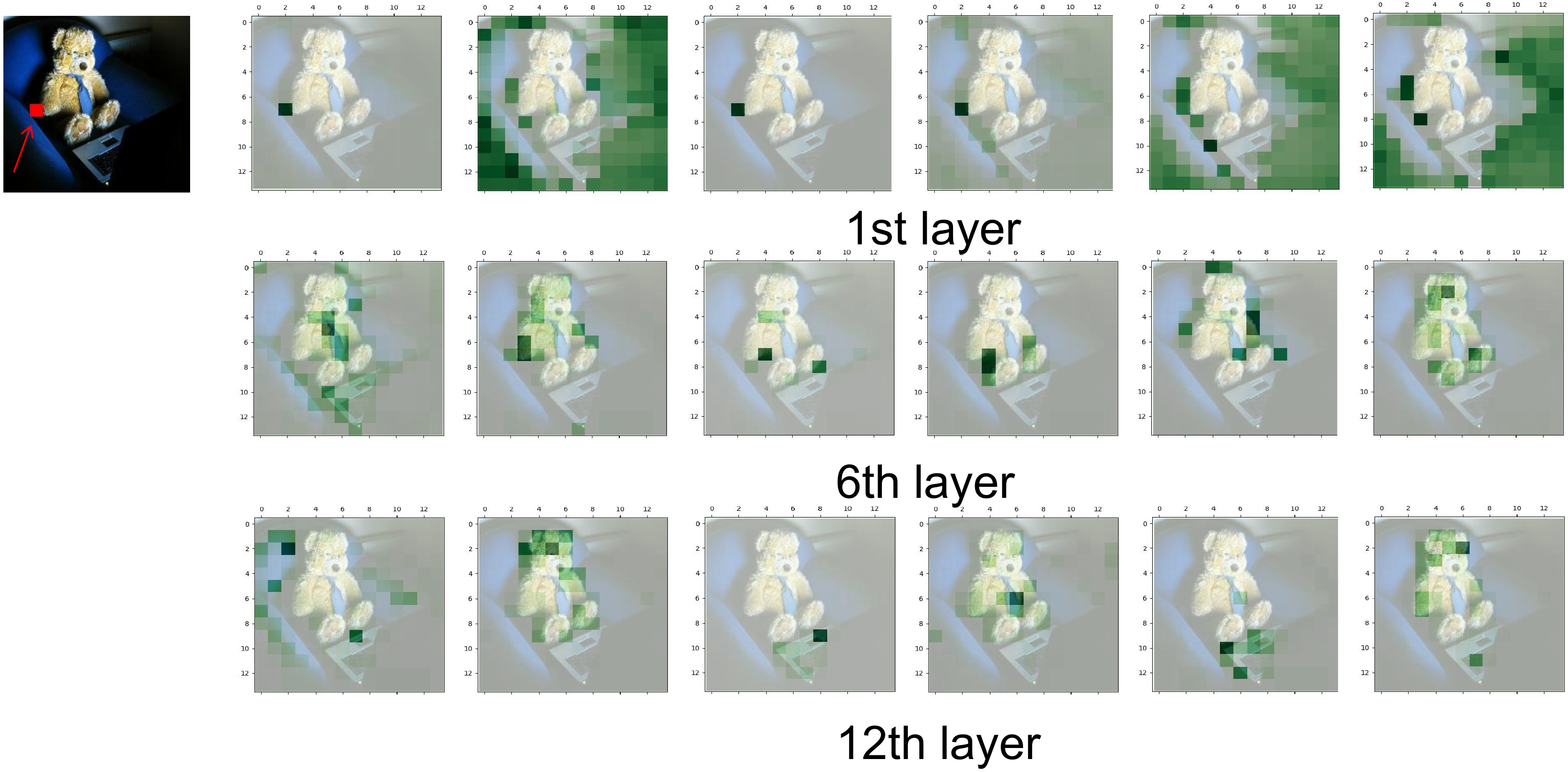}
\end{center}
   \vspace{-0.3cm}
   \caption{Visualization of the predicted encoder self-attention weights of different layers and attention heads. The image at the upper left corner is the raw image and the red point on it is the chosen query patch. The first, second and third row are the  attention weights visualization of the 1st, 6th, 12th encoder layer, respectively. The columns show different heads in given layers.}
\label{fig:encoder_att}
\end{figure}

\begin{figure}[h]
\begin{center}
\includegraphics[width=1.0\linewidth]{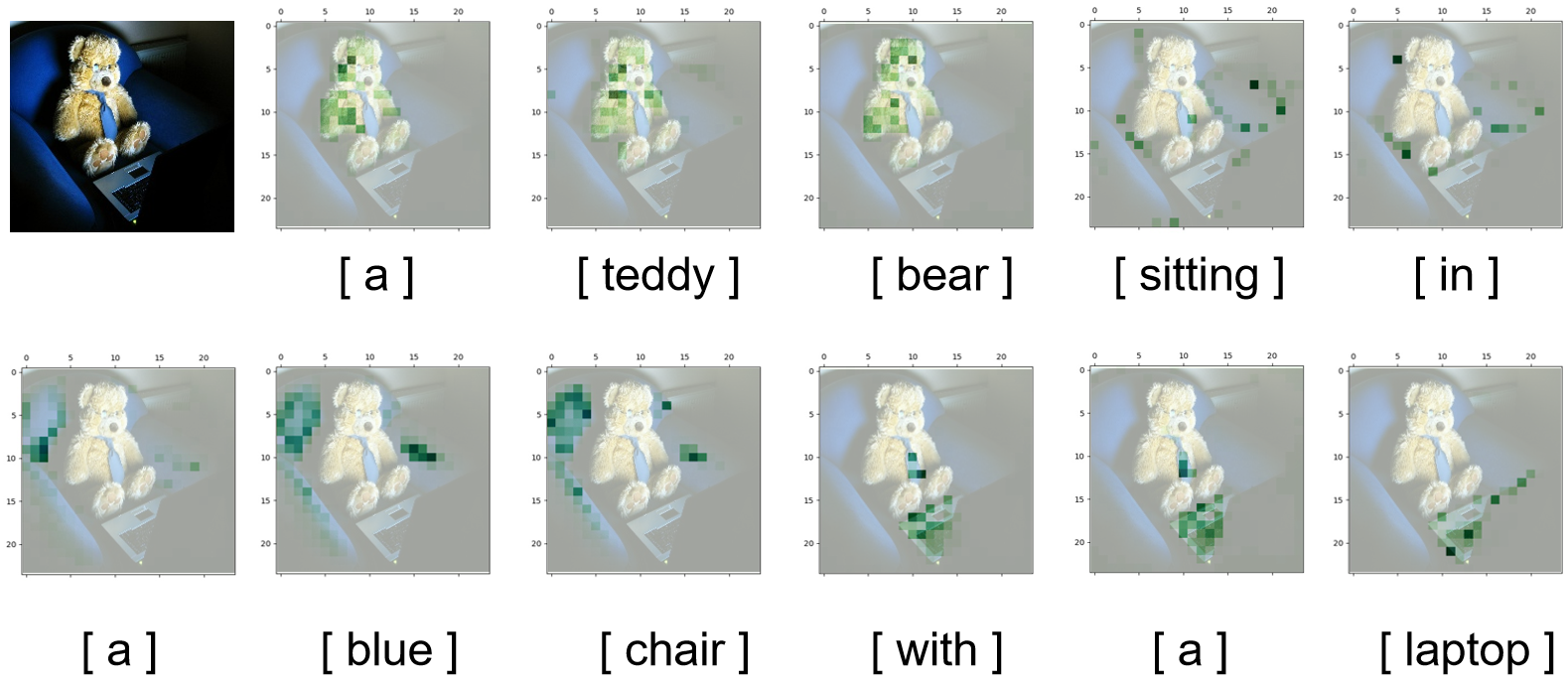}
\end{center}
   \vspace{-0.5cm}
   \caption{Visualization of the attention weights computed by the ``words-to-patches" cross attention in the last decoder layer. ``A teddy bear sitting in a blue chair with a laptop" is the caption generated by our model.}
   \vspace{-0.2cm}
\label{fig:decoder_att}
\end{figure}
\subsection{Attention Visualization}
In this section, we take one example image to show the caption predicted by  CPTR model and visualize both the self-attention weights of the patch sequences in the encoder and ``words-to-patches" cross attention weights in the decoder. With regards to the encoder self-attention, we choose an image patch to visualize its attention weights to all patches. As shown in  Fig. \ref{fig:encoder_att}, in the shallow layers, both the local and global contexts are exploited by different attention heads thanks to the full Transformer design which can not be fulfilled by the conventional CNN encoders. In the middle layer, model tends to pay attention to the primary object, i.e., ``teddy bear" in the image. The last layer fully utilizes global context and pays attention to all objects in the image, i.e., ``teddy bear", ``chair" and ``laptop". 

Besides, we visualize the ``words-to-patches" attention weights in the decoder during the caption generation process. As is shown in Fig. \ref{fig:decoder_att}, CPTR model can correctly attend to appropriate image patches when predicting every word.

\section{Conclusions}
\label{sec:typestyle}
In this paper, we rethink image captioning as a sequence-to-sequence prediction task and propose CPTR, a full Transformer model to replace the conventional ``CNN+Transformer" procedure. Our network is totally convolution-free and possesses the capacity of  modeling global context information at every layer of the encoder from the beginning. Evaluation results on the popular MS COCO dataset demonstrate the effectiveness of our method  and we surpass ``CNN+Transformer" networks. Detailed visualizations demonstrate that our model can exploit long range dependencies from the beginning and the decoder ``words-to-patches" attention  can precisely attend to the corresponding visual patches to predict words. 

\clearpage

\bibliographystyle{IEEEbib}
\bibliography{refs}
\end{document}